\documentclass[letterpaper, 10 pt, conference]{ieeeconf}  

\IEEEoverridecommandlockouts                              

\usepackage[backend=biber,style=ieee,citestyle=numeric-comp]{biblatex}
\usepackage{hyperref}
\usepackage{amsmath}
\usepackage{amssymb}

\usepackage{epsfig}
\usepackage{xcolor}
\usepackage{graphicx}
\usepackage{caption}
\usepackage{subcaption}
\usepackage{multirow}
\usepackage{booktabs}
\usepackage{cancel}

\title{\LARGE \bf
ColonMapper: topological mapping and localization for colonoscopy
}

\author{Javier Morlana, Juan D. Tard\'os and J.M.M. Montiel
\thanks{This work was supported by the EU-H2020 grant 863146: ENDOMAPPER, the Spanish government grants PID2021-127685NB-I00,  and by Arag\'on government grant DGA\_T45-17R.}
\thanks{The authors are with the Instituto de Investigaci\'on en Ingenier\'ia de Arag\'on (I3A), Universidad de Zaragoza, Mar\'ia de Luna 1, 50018 Zaragoza, Spain. E-mail: \{jmorlana,tardos,josemari\}@unizar.es.} }

\begin{document}

\maketitle

\begin{abstract}

We propose a topological mapping and localization system able to operate on real human colonoscopies, despite significant shape and illumination changes. The map is a graph where each node codes a colon location by a set of real images, while  edges represent traversability between nodes. 
For close-in-time images, where scene changes are minor, place recognition can be successfully managed with the recent transformers-based local feature matching algorithms. However, under long-term changes  --such as different colonoscopies of the same patient-- feature-based matching fails. 
To address this, we train on real colonoscopies a deep global descriptor achieving high recall with significant changes in the scene. The addition of a Bayesian filter boosts the accuracy of long-term place recognition, enabling relocalization in a previously built map. Our experiments show that ColonMapper is able to autonomously build a map and localize against it in two important use cases: localization within the same colonoscopy or within different colonoscopies of the same patient. Code: \href{https://github.com/jmorlana/ColonMapper}{github.com/jmorlana/ColonMapper}.


\end{abstract}

\section{Introduction}

\label{section:intro}
\noindent Simultaneous Localization And Mapping (SLAM) has a long tradition in out-of-the-body robotics and AR. The goal of SLAM is to localize a camera while building a map of an unknown environment. We can find two main approaches to solve the SLAM problem: \textit{metric} and \textit{topological}. Metric SLAM aims to estimate 6DoF camera position, while building a geometrical 3D map of the environment, that can be sparse or dense. On the other hand, topological SLAM avoids the metric problem, building a map that is just a graph whose nodes represent distinctive places connected by edges that model neighboring relations.


Colonoscopies are a tricky subject for SLAM due to two main challenges. The first is related to the variations in the images, that can suffer from extreme illumination changes, specularities and even different type of light (e.g. narrow-band imaging). Traditional image registration techniques fail in this scenario, and new lines of research involving neural networks are being investigated. The second is related to the variations in the environment itself. A colonoscopy involves deformations, fluids and surgical tools, making it really hard for metric SLAM that assumes static environments. A topological SLAM would presumably have an easier job, only having to consider relationships between images. 

\begin{figure}[t]
    \centering
    \includegraphics[width=\columnwidth]{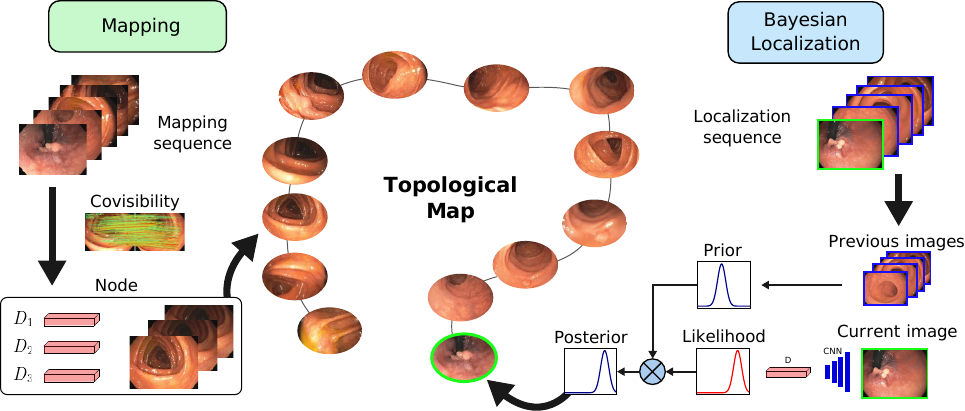}
    \caption{ColonMapper builds a map from a sequence, localizing another sequence against it. Our Bayesian filter driven by deep global descriptors enables intra and cross-sequence localization. }
    \vspace{-0.1cm}
    \label{fig:colonmapper}
\end{figure}

In this work, we focus on topological SLAM for the specific domain of colonoscopies. We build a topological map using neural networks to model the similarities between images (Fig.~\ref{fig:colonmapper}). In the end, a graph is obtained, where each node is formed by a set of different images observing the same place. Edges link places that are close to each other. We present two use cases for our system in real colonoscopy data: intra-sequence, where we build a map and localize within the same sequence, and cross-sequence, where we build a map of one sequence and localize a different sequence of the same patient. For localization we use deep global descriptors trained on real colonoscopies, and a Bayesian filtering approach that boosts the accuracy while keeping a high recall. The contributions of this work are:


\begin{itemize}
    \item We present  ColonMapper, a mapping and localization system able to process real colonoscopies. To the best of our knowledge, it is the first system able to create a map of the whole colon.
    \item We demonstrate our ability to detect covisible frames, by means of the local feature matcher LoFTR in the short-term and a deep global descriptor in the long-term. 
    \item We propose a novel strategy to mine hard-positive and hard-negative image pairs from the same colonoscopy. This forces the deep descriptor to focus on avoiding confusion within the same colonoscopy, boosting performance.
    \item An extensive evaluation in a colon phantom and in real colonoscopies, showing our exceptional capabilities for localizing in intra and cross-sequence scenarios.
\end{itemize}

\section{Related work}

\label{sec:related}
\noindent \textbf{Metric visual SLAM} methods are mature in out-of-the-body robotics and AR applications. Feature-based approaches \cite{mur2015orb,campos2021orb} estimate the sparse multi-view scene geometry via geometric bundle adjustment,  relying on discrete feature matching between images. Alternatively, direct methods \cite{engel2014lsd,engel2017direct}  recover scene geometry by means of photometric bundle adjustment relying on high-gradient pixels, without the need for feature detection.  Metric sparse VSLAM \cite{mur2015orb} has been brought to quasi-rigid laparoscopy in mid-size scenes to compute multi-view stereo dense maps in \cite{Mahmoud2019}.  SAGE-SLAM \cite{liu2022sage} builds a SLAM pipeline with learned depth and features, producing dense maps and accurate camera poses for the specific environment of rigid endoscopic endonasal surgery.  CNN-SLAM \cite{Tateno_2017_CVPR} proposed to combine LSD-SLAM \cite{engel2014lsd} with single depth predictions from a CNN \cite{laina2016deeper} to achieve dense monocular SLAM in out-of-the-body scenes. It has been brought to colonoscopy in RNN-SLAM  \cite{ma2019real,ma2021rnnslam} replacing  LSD-SLAM by the better performing DSO \cite{engel2017direct}. It has been proven able to process disjoint chunks of colonoscopy producing dense maps, being nowadays the best performing dense SLAM system in real colonoscopy. Despite not being a SLAM system, it is worth mentioning the feature-based SfM COLMAP \cite{schonberger2016structure} because it has proven  to compute sparse 3D geometry from close-in-time real colonoscopy frames. This method is not sequential and is quite time-consuming, but it can produce valuable 3D information in colonoscopy where no other sensor or medical image modality can. This pseudo ground truth geometry has been exploited to train the single-view depth network in RNN-SLAM \cite{ma2019real}.

\vspace{0.1cm}

\noindent \textbf{Topological visual SLAM} algorithms have a great potential in endoscopy where track losses are frequent due to occlusions, motion blur, deformations or sudden camera motion. They avoid the burden of recovering scene geometry and only focus on recognizing the visited locations by their visual appearance. Traditional methods for topological SLAM \cite{cummins2008fab,cummins2011appearance,angeli2008incremental} relied on features like SIFT \cite{lowe2004sift}. They treat loop-closure as an image retrieval problem, where the algorithm looks for an image in the map similar to the current one by means of a Bag-of-Words representation of the visual appearance of places, further verified by a rigid geometry validation. These ideas were extended to binary ORB descriptors in \cite{galvez2011real}. A Bayesian filter is proposed in \cite{angeli2008fast} to improve accuracy of loop closure detection. A similar approach is proposed in \cite{Pueyo22} for colonoscopies, using AKAZE \cite{alcantarilla2013fast} keypoints.

\vspace{0.1cm}

\noindent \textbf{Visual recognition} has also evolved with the deep learning era, and multiple works \cite{arandjelovic2016netvlad,radenovic2018fine} have proposed neural networks for the task of retrieval. Place recognition is being brought to colonoscopies with works like \cite{morlana2021self,ma2021colon10k} fine-tuning neural networks with colonoscopy data, leveraging on SfM  \cite{schonberger2016structure} to detect covisible images in real colonoscopies among close-in-time frames. Deep learning has also provided unprecedented performance in the local features matching task, reducing the necessity of rigid geometry verification to achieve an acceptable precision with the detector SuperPoint \cite{detone2018superpoint} combined with the matcher SuperGlue \cite{sarlin2020superglue} or the recent transformer-based detector-free matcher LoFTR \cite{sun2021loftr}.

Our mapping and localization builds on several works: 
LoFTR \cite{sun2021loftr} to detect close-in-time covisible frames, without needing to apply the rigid geometry verification, coping with the small deformations prevalent in consecutive colonoscopy frames; GeM/NetVLAD descriptors \cite{radenovic2018fine, arandjelovic2016netvlad} for building our image retrieval network \cite{berton2022deep}, that having been trained with close-in-time SfM covisible frames, successfully identify reobservations of distant frames; and Bayesian filtering \cite{angeli2008incremental,angeli2008fast} to boosts localization accuracy. Thanks to the topological approach and the use of learned features, we are able to build meaningful maps of the complete colon and localize in them, even in different colonoscopies of the same patient.


\section{ColonMapper}

ColonMapper works in two steps, \textit{topological mapping} and \textit{Bayesian localization}, both driven by two deep features:

\vspace{0.1cm}

\noindent \textbf{Global descriptor}: following \cite{arandjelovic2016netvlad,morlana2021self}, we obtain a global descriptor for every processed image. We employ a R50 architecture with the final fully-connected layers removed, followed by a GeM \cite{radenovic2018fine} or NetVLAD agreggation technique \cite{arandjelovic2016netvlad} and L2-normalization. For each image $I_{i}$, we obtain a descriptor $\mathbf{d_i}$  that can be compared to others by dot product, resulting in a real-valued similarity score $ s \in \left [ 0, 1 \right ]$. It is used in the \textit{node creation} in the topological mapping (Sec. \ref{subsec:topological}) and in the \textit{likelihood} estimation in the Bayesian localization (Sec. \ref{subsec:loopclosure}). Our network is trained in real colonoscopy data using the framework proposed by \cite{berton2022deep} (Sec. \ref{subsec:implementation}).

\vspace{0.1cm}

\noindent \textbf{LoFTR} \cite{sun2021loftr} is employed as our matching method due to its capacity to yield dense matches in low-textured regions without requiring a separate step for keypoint detection. LoFTR extracts dense features for both images simultaneously by leveraging self and cross attention layers in transformers, enabling it to search for correspondences at a coarse level. This is particularly valuable in medical imaging contexts, where local features may lack discriminative power due to low scene texture, but global relationships between images can still be identified. We use LoFTR in the \textit{node creation} step in the topological mapping (Sec. \ref{subsec:topological}).

\subsection{Topological mapping for colonoscopies}
\label{subsec:topological}


Our topological map is a graph composed by nodes that represent \textit{places} (distinctive sections of the colon anatomy), and edges that link places that are connected in space. 
A node is defined as a set of images that observe the same place in the colon. Every image has an associated global descriptor. These images should be different from each other in order to have meaningful views of the place, so we should avoid the insertion of too similar frames. We should also be able to automatically detect and discard non-distinctive places, which can be originated from walls and fluids in typical colonoscopies.

Our mapping algorithm starts with the \textit{node creation} step. It reads a new image and inserts it along with its associated global descriptor into a new node $N_{new}$. We then continue reading images, skipping them if their similarity with the last image added to the node is greater than $s_{skip}$. When a new coming image is different enough or $n_{skip}$ frames has been skipped, we run a matching step using LoFTR between the incoming image and the last image added to the node. If the matching is successful, the image is included with its global descriptor in $N_{new}$. If tha matching fails, we close $N_{new}$ and, if it has at least 3 images, we insert it in the map linked to the last inserted node. Otherwise, we discard $N_{new}$ to avoid nodes depicting walls or low-visibility places. We start another new node with the next incoming image.


\subsection{Bayesian Localization}
\label{subsec:loopclosure}





Our localization pipeline receives images sequentially, and computes a probability distribution of the localization of the current image $I_t$ against all the nodes in the map with a Bayesian filtering approach.  
The localization hypotheses at time $t$ are represented by the distribution $p\left ( S_t = i | z^t \right )$ for $i \in \{ 0,\ldots,n \}$ where $n$ is the number of nodes  in the map and $z^t = \{ z_0,\ldots,z_t \}$ is the set of observations until instant $t$, where $z_j$ is the global descriptor associated with image $I_j$. The\textit{ full posterior}  
$p\left ( S_t | z^t \right )$ for the localization at time $t$ is computed using the classical prediction and update equations of a Bayesian localization filter \cite{thrun2005probabilistic}: 
\begin{eqnarray}
    p\left ( S_t | z^{t-1} \right ) &=&  \sum_{j=1}^{n} p\left ( S_t | S_{t-1} = j \right ) p\left ( S_{t-1} = j | z^{t-1} \right ) \label{eq:prediction}  \\
    p\left ( S_t | z^t \right ) &=& \eta \; p\left ( z_t | S_t \right ) \; p\left ( S_t | z^{t-1} \right ) \label{eq:posterior}
\end{eqnarray}




Colonoscopy priors are added into the system through our \textit{time evolution model} $p\left ( S_t | S_{t-1} = j \right )$  that is applied to the previous estimation. It considers that colonoscopies are linear, so it is likely to jump to close nodes in the map (within a distance $m$), although there is a small probability $\alpha$ of jumping to a more distant node:

\begin{equation}
p\left( S_t = i | S_{t-1}=j\right) = \begin{cases}
\dfrac{1-\alpha}{2m+1} & \text{if }  |i-j| \leq m \\[10pt]
\dfrac{\alpha}{n-2m-1} & \text{otherwise}  \end{cases} 
\end{equation}

In practice, $\alpha$ works as a \textit{smoothing} parameter that avoids the system to be too confident in its localization, specially  after camera occlusions. 



Our \textit{likelihood} $p\left ( z_t | S_t=i \right )$ is detailed next. We compare the global descriptor from 
$I_t$ against all the images for the node $N_i$ in the map and select the maximum, obtaining a single score relating $I_t$ and $N_i$. Once the scores with respect to each node in the map have been obtained, we retain the top-7 scores and set the rest to a value of 0.2. 


In the filter update phase (Eq.\ref{eq:posterior}), the likelihood is composed with the prediction to obtain the full posterior. Finding the posterior mode is not enough for robust localization, as difficult images can degrade the distribution and, if nodes are similar, the probability may not show a single peak but be diffused over neighboring nodes. As in \cite{angeli2008fast}, we take profit of the similarity between neighboring nodes by summing the probability of every node $N_i$ and its neighbors within distance $w$ , obtaining $p_{sum}$. We will accept a localization for image $I_{t}$ if $p_{sum} > p_{sum,th}$ for any node $N_{i}$ in the map.

\vspace{0.1cm}

\noindent \textbf{Rejecting spurious observations:} colon walls and low-visibility areas in real colonoscopies can degrade the performance of the localization by adding noise as these images are not visually recognizable. To solve this issue, we create a \textit{reject node} by manually selecting confusing frames from an unseen sequence. The \textit{reject node} has 25 images depicting walls and fluids. Our algorithm rejects the new image if the average score of the top-3 map nodes is lower than the average of the top-3 images in the \textit{reject node}. In this case, no localization is accepted, and only the prediction phase of the Bayesian filter is performed  (Eq.\ref{eq:prediction}).


\section{Experiments}




We first train and test the deep global descriptors for the task of image retrieval in real colonoscopies, and then analyze the performance of our Bayesian localization method in phantom and real colonoscopies. 

\subsection{Deep global descriptor}
\label{subsec:implementation}

\noindent \textbf{Training}: following \cite{morlana2021self,ma2021colon10k}, we train an image retrieval network on colonoscopy data for our global descriptors. We use 20 sequences from the Endomapper dataset \cite{azagra2022endomapper}: 18 for training and 2 for validation.  We train using image triplets, so we need to obtain covisibility information to create pairs of positive (covisible) and negative (non-covisible) images. Covisibility was found in two ways. First, we processed our sequences with COLMAP \cite{schonberger2016structure} in sequential mode, obtaining positive pairs from images reconstructed in the same COLMAP cluster that share at least one 3D point, resulting in 82k training and 15k validation images. From the training images, we randomly selected 28k queries that form triplets during training. Secondly, we manually labelled clusters that were covisible, but COLMAP was not able to merge. We manually labelled additional positive pairs by randomly sampling pairs coming from the same COLMAP cluster or from two different covisible clusters. These pairs were manually inspected, keeping the covisible ones. This second stage adds 1.3k \textit{manual pairs} that COLMAP was not able to find. Manually labeling covisible clusters allow us to mine negatives from the same sequence, differently from other works \cite{morlana2021self, ma2021colon10k} that mine negatives strictly from different sequences. We believe that same-sequence negatives are of paramount importance, as in practice, you would never need to disambiguate between two sequences of different patients. Besides, two images from the same colon region of different patients can look very similar, hindering the network's convergence. 

Training is done with the triplet loss using one positive and ten negatives per every query, using a modified version of the image retrieval framework proposed by \cite{berton2022deep}. We sample 5000 queries per epoch, re-mining hard-positives and hard-negatives every 1000 queries. The pool of negatives is set to 5000, while every query frame has a pool of ten random positives among all the ones found by COLMAP.  We have three strategies to chose one positive among the pool of ten: \textit{easy} uses the most similar according to the global descriptor, \textit{semi-hard} uses the positive that is ranked in the middle and \textit{hard} uses always the most difficult, the furthest in descriptor space. For the case of the \textit{manual pairs}, none of this apply, as each query only has one manually labelled positive, that is used all the time. Training  is performed for at most 30 epochs with a patience of 5 epochs. The best checkpoint is selected based on the mean Average Precision (mAP) \cite{philbin2007object} in the validation set as in \cite{radenovic2018fine}.

\begin{figure*}[t]
    \centering
    \includegraphics[width=\textwidth]{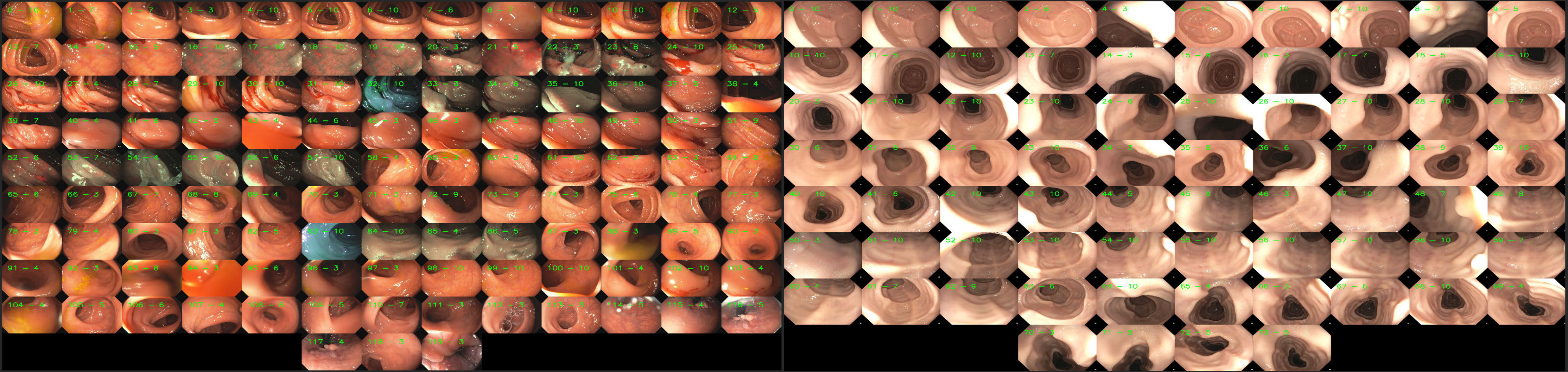}
    \caption{Topological maps from the withdrawal phase of \textit{Seq\_027} in Endomapper (left) and from \textit{seq1} in C3VD (right). }
    \vspace{0.1cm}
    \label{fig:maps}
\end{figure*}

\begin{figure*}[t]
    \centering
    \includegraphics[width=0.95\textwidth]{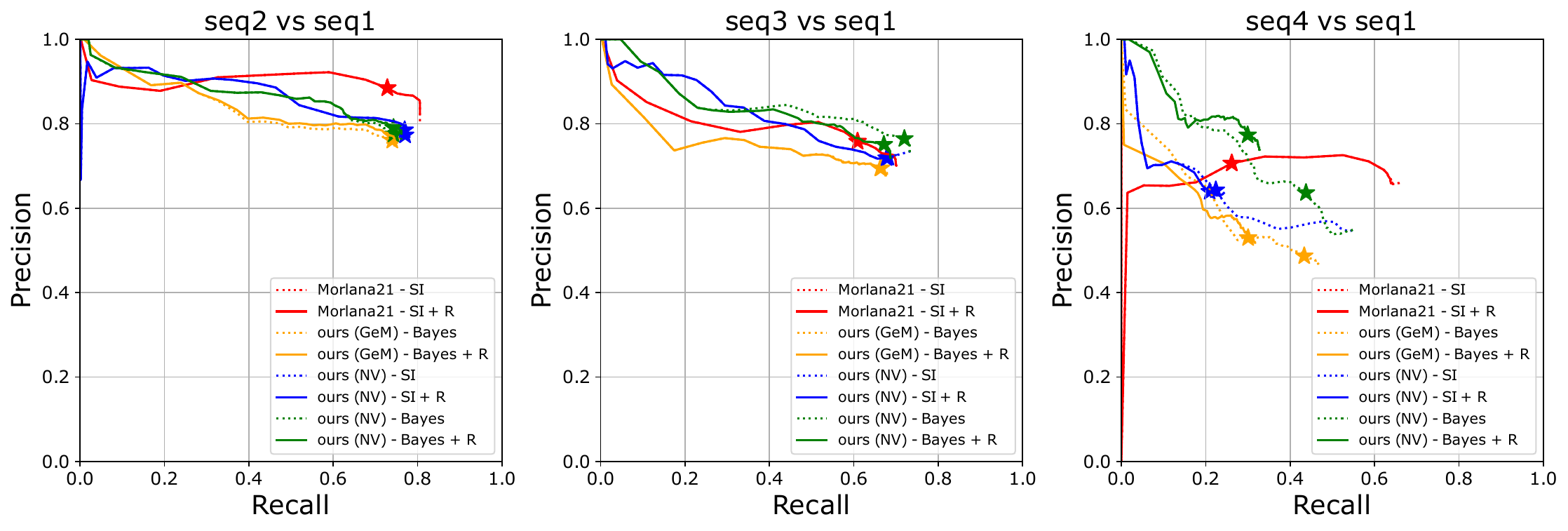}
    \vspace{-0.1cm}
    \caption{Precision-Recall curves for C3VD. Categories: Single-Image (SI), Bayesian (Bayes), Reject spurious (R). }
    \label{fig:PR-C3VD}
\end{figure*}

\vspace{0.1cm}

\noindent \textbf{Evaluation}: we report several models in Table \ref{table:IR}, all of them using a ResNet-50 architecture, that gives us a good trade-off between capacity and performance. All models are pretrained on ImageNet and fine-tuned in our colonoscopy data. Layer 1 in ResNet is frozen during fine-tuning, while the other layers are optimized. The aggregators GeM \cite{radenovic2018fine} or NetVLAD \cite{arandjelovic2016netvlad} are set after layer4, as they produce better results according to \cite{berton2022deep}. Our work presents several differences with respect to our previous work \cite{morlana2021self}, that reported a model fully trained where the GeM layer was placed after the last layer (layer 5), was trained with a much smaller dataset (3 sequences), without performing hard-positive mining and mining negatives only from different sequences. As we show later, this training is very limited for real use case scenarios.

We target the network with the best capability to find long-term reobservation. To find this model, we created a test dataset by processing with COLMAP \textit{Seq\_027} and \textit{Seq\_035} from Endomapper \cite{azagra2022endomapper}, two colonoscopies of the same patient performed two weeks apart. We extracted 3 images from each of the COLMAP clusters obtained and use them as queries, giving a total of 84 queries for \textit{Seq\_027} and 78 queries for \textit{Seq\_035}. We did two experiments: \textit{intra} and \textit{cross-sequence} retrieval. We manually labelled, for each of the queries, all the covisible images that were reconstructed in a different COLMAP cluster than the one the query was extracted from. For \textit{intra}, we look for covisibility between queries and clusters from \textit{Seq\_027}, while for \textit{cross}, we do it between \textit{Seq\_035} queries and \textit{Seq\_027} clusters. Table \ref{table:IR} shows the mAP of the different models  and compares against \cite{morlana2021self}, showing a huge increase in all cases, due to the better training procedure (hard-positives mining and mining negatives from the same sequence) and the bigger amount of data. 

\begin{table}[t]
\resizebox{\columnwidth}{!}{%
    \centering
    \begin{tabular}{cccccccc}
        \toprule
        \multirow{2}{*}{Model} & \multirow{2}{*}{Backbone} & \multirow{2}{*}{Aggregation}  & Positive & mAP & mAP & mAP\\ 
                               &                             &                           & mining   & intra & cross & sum \\
        \midrule
        Morlana21 \cite{morlana2021self}         	 & R101  & GeM  & -  & 53.7 & 21.7 & 75.4 \\
         \midrule
        R50-GeM-E    				 & R50  & GeM  & Easy  & 75.0 & 30.1 & 105.1 \\
        R50-GeM-SH    	             & R50  & GeM  & Semi-hard  & 76.7 & 33.8 & 110.5 \\
        \textbf{R50-GeM-H}   		 & R50  & GeM  & Hard & 73.3 & 38.5 & \textbf{111.8} \\
        \midrule
        R50-NV-E    	             & R50  & NetVLAD  & Easy  & 65.3 & 28.0 & 93.3 \\
        R50-NV-SH    				 & R50  & NetVLAD  & Semi-hard  & 73.8 & 33.1 & 106.9 \\
        \textbf{R50-NV-H}   	     & R50  & NetVLAD  & Hard  & 75.7 & 33.0 & \textbf{108.7} \\

        \bottomrule
    \end{tabular}}
    \vspace{-0.1cm}
    \caption{Image Retrieval analysis. }
    \label{table:IR}
    \vspace{-0.2cm}
\end{table}

\subsection{Localization performance }

\noindent \textbf{Mapping and localization tuning}: for LoFTR
\cite{sun2021loftr}, we use the off-the-shelf implementation from Kornia \cite{riba2020kornia} with outdoor weights. We preprocessed a mask in order to avoid matches in the black corners of our images. Matching is successful if it yields more than 100 matches. In the mapping phase, we use $s_{skip} = 0.6$ and $t_{skip} = 7$ for Endomapper \cite{azagra2022endomapper}, while $t_{skip} = 10$ for C3VD \cite{bobrow2023colonoscopy}. Nodes can have at most 10 images before starting a new node. For localization, $\alpha=0.05$, $m=2$ and $w=3$. In the likelihood, top-7 scores lower than 0.5 (0.9) are set to 0.3 (0.5) for NetVLAD (GeM), reducing the impact of false positives in the posterior probability.

\vspace{0.1cm}

\noindent \textbf{Evaluation:} we evaluate our localization capabilities in phantom and real colonoscopy data. The evaluation procedure starts by building what we call the \textit{canonical maps} (Fig. \ref{fig:maps}), that are used for localization by all the models evaluated. \textit{Canonical maps} are needed to have a common set of places to localize, as the maps can slightly change for every global descriptor. We choose to build \textit{canonical maps} in C3VD (Sect. \ref{subsubsec:c3vd}) and Endomapper (Sect. \ref{subsubsec:endomapper}) with model R50-NV-H, as we observed better performance in our initial evaluations. For the model proposed by Morlana et al. \cite{morlana2021self} and for R50-GeM-H, we simply extract new global descriptors for the images included in the map. All nodes in the canonical map are assigned with a ground truth (GT) label, that represents the colon region where it is located. Labels are: \textit{retroflexure, rectum, sigmoid, descending, transverse, ascending} and \textit{cecum}. The canonical map is built from a sequence different than the ones used for localization.  

To have manageable data to evaluate, we process 1 out of 5 frames sequentially from the localization sequences, skipping the frames in between. Each of the frames is assigned with one of the GT labels used for the map, plus the label \textit{none}, which is assigned to those frames that can not be localized, either because they are not visually recognizable or because we weren't sure where they belong. The labeling procedure was done manually by ourselves. For labeling Endomapper sequences (Sect. \ref{subsubsec:endomapper}) we had the help of the real-time footage provided by the doctors when the sequences were recorded, that roughly states where the colonoscope is. 

\vspace{0.1cm}

\noindent \textbf{Baselines:} we compare the Single-Image approach proposed by Morlana et al. \cite{morlana2021self} against our models R50-NV-H and R50-GeM-H, in the categories Single-Image (SI) or Bayesian (Bayes), with the possibility of Rejecting (R) spurious observations. For Single-Image methods, the accepted localization is the map node with highest similarity $s$. For SI + Reject, the system can Reject frames if the highest $s$ comes from the \textit{reject node}. For Bayes and Bayes + R, we select the localization with $p_{sum}$ as explained in Sect. \ref{subsec:loopclosure}.

\vspace{0.1cm}

\begin{figure*}[t]
    \centering
    \includegraphics[width=0.95\textwidth]{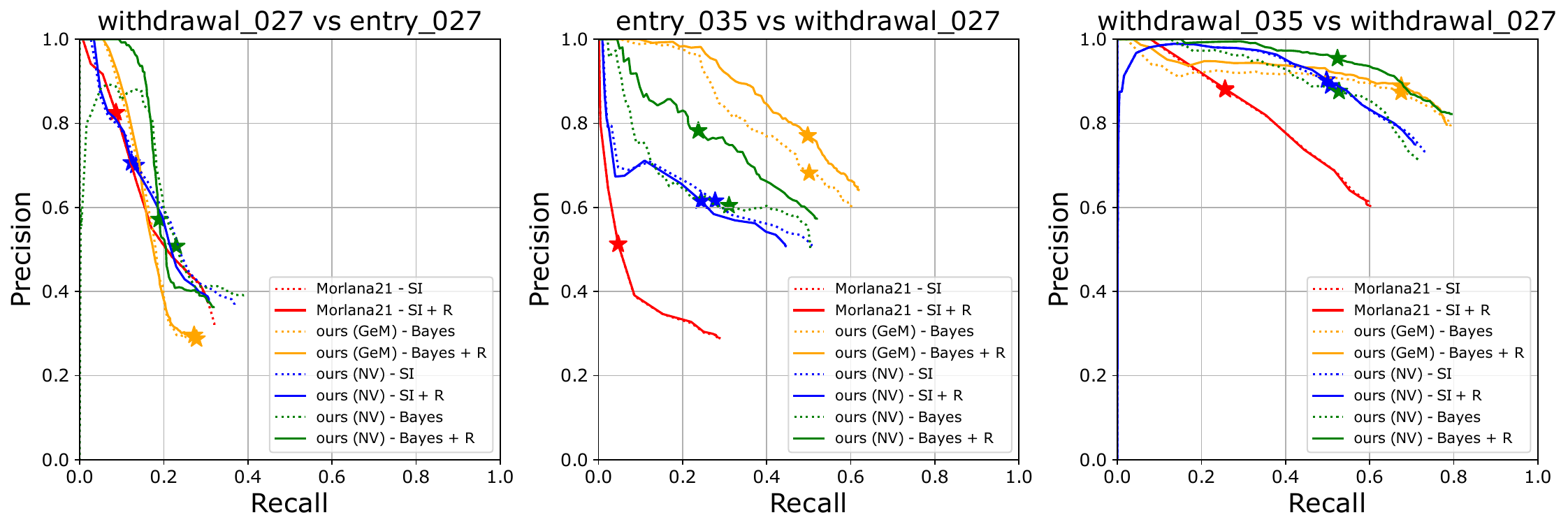}
    \vspace{-0.2cm}
    \caption{Precision-Recall curves for Endomapper. Categories: Single-Image (SI), Bayesian (Bayes), Reject spurious (R).}
    \label{fig:PR-endomapper}
\end{figure*}

\noindent \textbf{Metrics:} we show Precision-Recall curves for comparison. As every image is localized depending on a probability/score threshold, the curves represent precision and recall for all the possible values of an acceptance threshold. For Single-Image, this threshold corresponds to the similarity score $s$, while for the Bayesian approach, the threshold $p_{sum,th}$ is defined w.r.t. to the summed probability $p_{sum}$. The localization is correct if the GT label of the estimated node matches the GT label of the located frame. Thus, precision $\mathbf{P}$ and recall $\mathbf{R}$ are defined as follows:
\begin{equation}
  \mathbf{P} =\frac{TP}{Retrieved}\;\; , \;\; \mathbf{R} =\frac{TP}{Relevant}
  \label{eq:eq_PR}
\end{equation}
where $TP$ corresponds to all the correctly localized frames, $Retrieved$ is the number of frames where we accept a localization 
 and $Relevant$ corresponds to all the frames with a GT label. Frames labelled as \textit{none} are excluded from the evaluation when estimating $\mathbf{P}$ and $\mathbf{R}$, but they are still processed by our algorithms as in a real in-vivo sequence. Thresholds for the operating points appearing as a star in \ref{fig:PR-C3VD} and \ref{fig:PR-endomapper} are selected to have a good trade-off between $\mathbf{P}$ and $\mathbf{R}$ in all datasets. Acceptance thresholds for Single-Image approaches are $s_{GeM}=0.85$ and $s_{NV}=0.55$, while for Bayesian, $p_{sum,th}=0.5$. 

\subsubsection{\textbf{Colonoscopy 3D Video Dataset (C3VD)} \cite{bobrow2023colonoscopy}}
\label{subsubsec:c3vd}

 provides 4 screening sequences recording a 3D-printed phantom colon, with $\approx$~5k frames each. Starting at the cecum, a full trajectory is recorded while a doctor withdraws the colonoscope. Each of the screening videos show the same 3D model but with a different texture, which could simulate the colon of the same patient recorded in 4 different moments. C3VD allows us to evaluate our Bayesian localization in a controlled setting. Typical challenges such as deformations, surgical tools or fluids, do not appear in these sequences, making them easier that real colonoscopies. 

\vspace{0.1cm}

\noindent \textbf{Experiments}: the \textit{canonical map} is obtained for the screening \textit{seq1} (Fig. \ref{fig:maps}). It covers the whole phantom colon from cecum to sigmoid. Sequences \textit{seq2}, \textit{seq3} and \textit{seq4} are localized against \textit{seq1} map.

\vspace{0.1cm}

\noindent \textbf{Results:} the PR-curves can be seen in Fig. \ref{fig:PR-C3VD}. \textit{seq2} is the most similar to \textit{seq1}, being both low-textured and light-pink colored. On the contrary, \textit{seq3} and \textit{seq4} have richer texture and a darker-pink color, being harder to localize than \textit{seq2}. Overall, all approaches rank high in \textit{seq2}, with \cite{morlana2021self} performing a bit better.  Interestingly, we can clearly see how NetVLAD + Bayes improves w.r.t. to the other approaches when the difficulty gets harder. The NetVLAD model also outperforms the one using GeM. In C3VD, the Reject node has little effect as it is a very controlled setting with almost no occlusions. We didn't label any frame as \textit{none} for C3VD. 

\subsubsection{\textbf{Endomapper dataset} \cite{azagra2022endomapper}}
\label{subsubsec:endomapper}

\begin{figure}[t]
    \centering
    \includegraphics[width=\columnwidth]{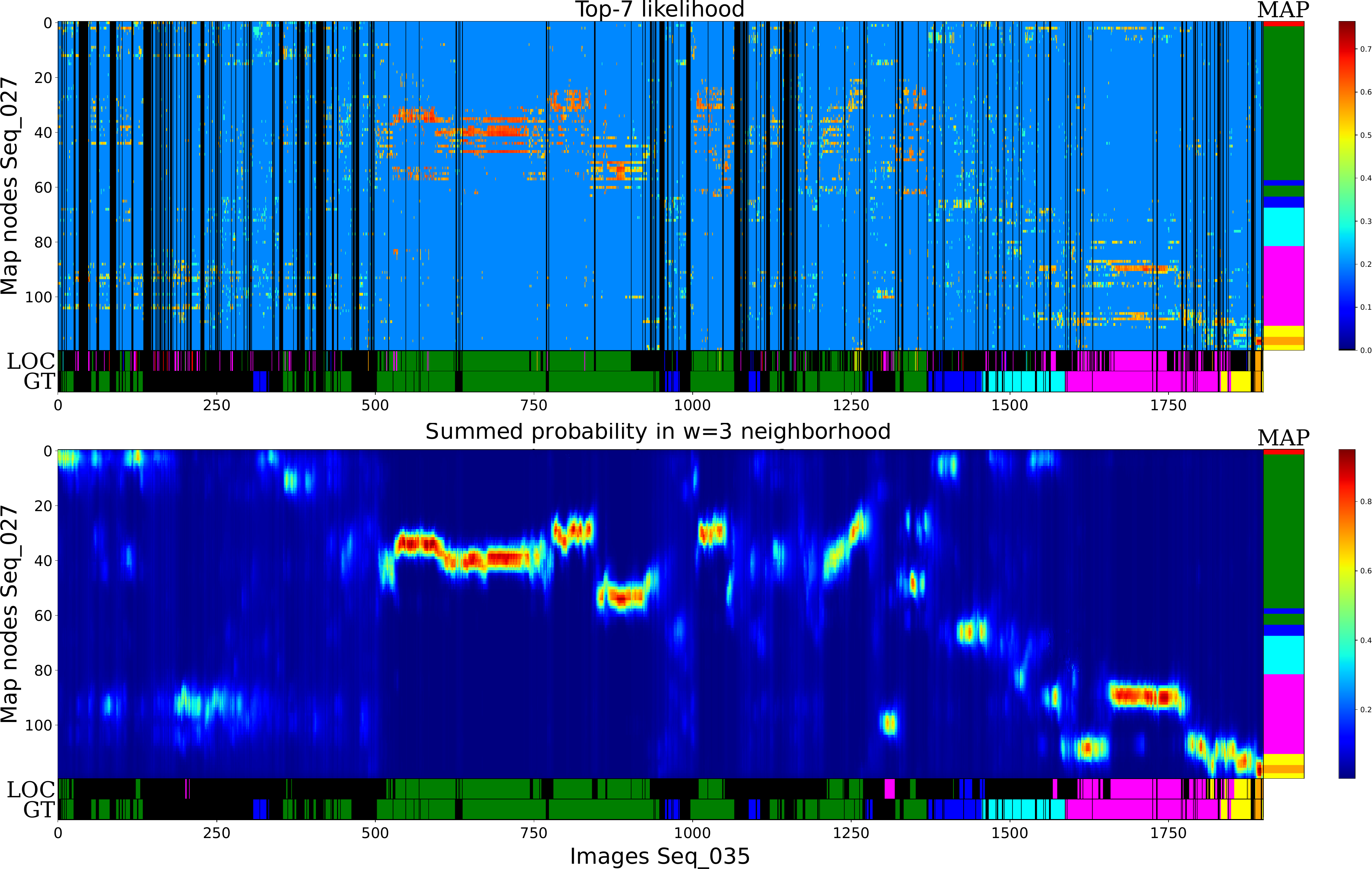}
    \caption{Likelihood (top) and $p_{sum}$ (bottom) for R50-NV-H - Bayes + R in  withdrawal\_035 vs withdrawal\_map\_027. Labels: \textit{\textcolor{red}{cecum}, \textcolor{green}{ascending}, \textcolor{blue}{transverse}, \textcolor{cyan}{descending}, \textcolor{magenta}{sigmoid}, \textcolor{yellow}{rectum}, \textcolor{orange}{retroflexure}, none}. }
    \label{fig:prob_evolution}
\end{figure}

 comprises 96 high quality videos of real colonoscopies. Differently from C3VD, Endomapper sequences are real medical procedures subject to challenges like strong deformations, illumination changes, surgical tools and fluids, being a much harder scenario for any mapping and localization system. As in Sec. \ref{subsec:implementation}, we use \textit{Seq\_027} and \textit{Seq\_035}, two explorations of the same patient, to prove the ability of ColonMapper to localize \textit{intra} and \textit{cross-sequence}.

\vspace{0.1cm}

\noindent \textbf{Experiments}: as \textit{canonical maps}, we built from \textit{Seq\_027} an \textit{entry\_map} from rectum to cecum and a  \textit{withdrawal\_map} from cecum to rectum (Fig. \ref{fig:maps}). We  run three experiments to show our intra and cross-sequence localization capabilities. For intra, we localize the withdrawal phase of \textit{Seq\_027} against the \textit{entry\_map} of \textit{Seq\_027}. For cross, we localize both the entry and the withdrawal of \textit{Seq\_035} against the \textit{withdrawal\_map} of \textit{Seq\_027}.



\vspace{0.1cm}

\noindent \textbf{Results:} PR-curves can be seen in Fig. \ref{fig:PR-endomapper}. In real data, we can see how Bayesian localization improves the Single-Image (SI) approach in all cases, and the Reject node boosts the Bayesian approach every time. Reject node is really useful in real sequences due to the appearance of walls and fluids that can degrade the Bayesian posterior, but has no effect in SI as \textit{none}-labeled frames, although badly localized, are not included in the evaluation and cannot influence future localizations. We can observe that intra-sequence results are worse than those obtained for cross-sequence, due to the inherent nature of colonocopies. In the entry phase, the colonoscope is inserted as fast as possible, trying to reach the cecum as soon as the doctor is able to. Withdrawals are much cleaner and careful, as the doctor wants to observe everything, being more amenable for mapping and localization.  

\begin{figure}[t]
    \centering    \includegraphics[width=\columnwidth]{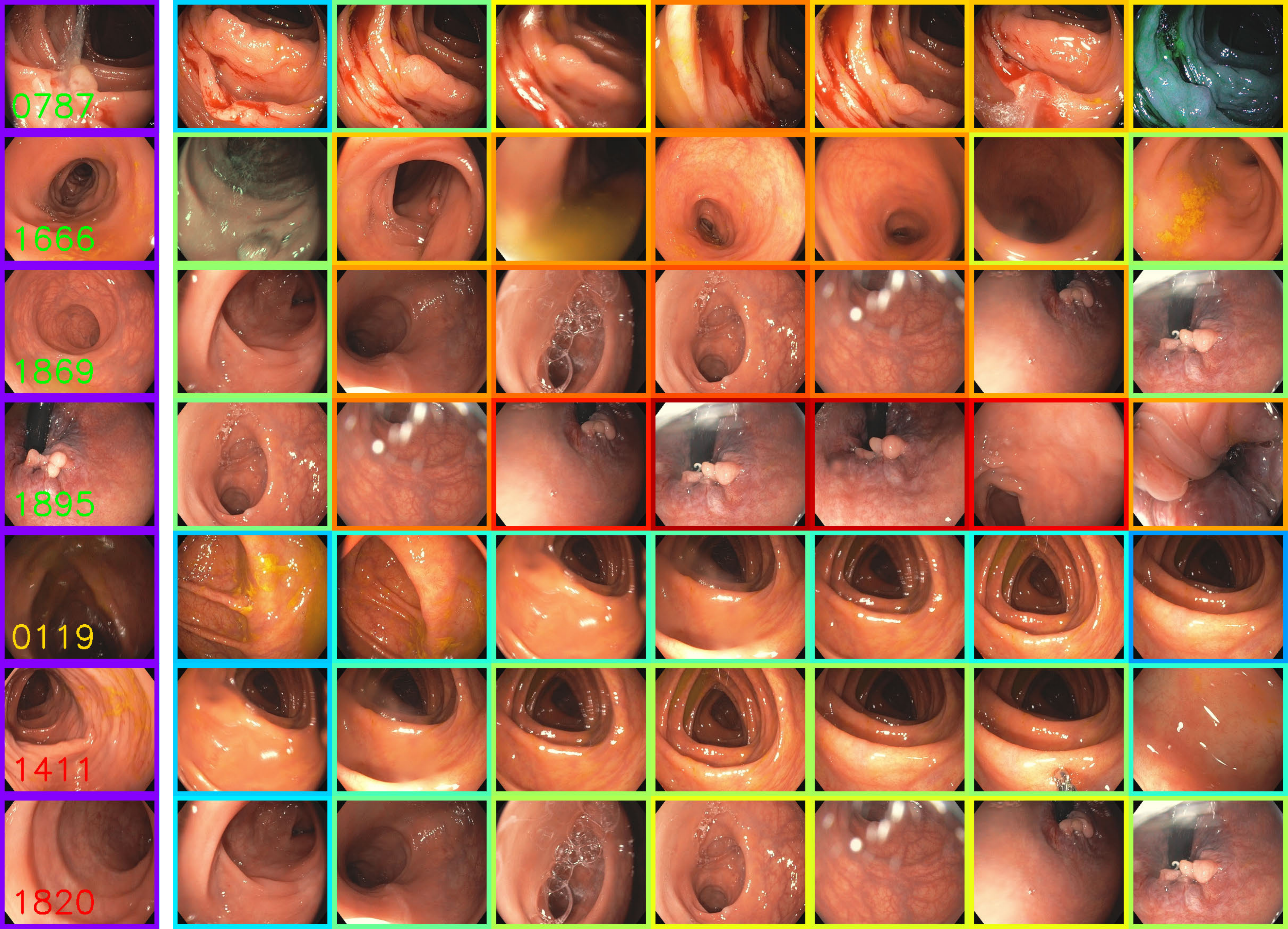}
    \caption{Localizations found by R50-NV-H with Bayes + R. }
    \label{fig:localizations}
    \vspace{-0.1cm}
\end{figure}

Comparing our models R50-GeM-H and R50-NV-H, we finally select the one with NetVLAD as it outperforms GeM in all cases except for entry\_035 vs withdrawal\_map\_027 in Fig. \ref{fig:PR-endomapper}. For the R50-NV-H with Bayes + R, we show in Fig. \ref{fig:prob_evolution} the top-7 likelihood and the evolution of the summed probability for the case of withdrawal\_035 vs withdrawal\_027. The rejected images are shown as black lines in the likelihood, that translates to a smoothing in the summed probability showed in the bottom. Below each of the plots, we can see the ground truth (GT) labels assigned to every image  and the localizations (LOC) obtained using the operating points for Single-Image (equivalent to use only the likelihood) and for Bayesian with Reject. The Bayesian approach is clearly more robust than the Single-Image approach, that jumps between regions too easily. The trajectory followed by the colonoscope can be observed, spending a lot of time in the ascending, then to the transverse, and finally to sigmoid, rectum and performing a retroflexure.

Some localization examples for this model are shown  in Fig. \ref{fig:localizations}. The localization queries are shown in the left column, while the window of 7 nodes around the one with highest summed probability are shown in the right. The first four rows show correct localizations whose $p_{sum} > 0.5$, showing remarkable capabilities of localizing places along the colon. The fifth row shows an example where $p_{sum} < 0.5$, so it was not localized. The last two rows show cases when the algorithm failed and selected an erroneous place, which can be understood due to the similarity of the images.

\section{Conclusions}

We have presented ColonMapper, the first mapping and localization system able to process the whole colon. Powered by deep neural networks, it is able to select different frames to build nodes that characterize colon places. More interestingly, we have shown that long-term localization is possible in colon regions of the same patient in different colonoscopies, thanks to a Bayesian filtering approach, which boosts the recall and precision of single-image deep visual place recognition. We see ColonMapper as a first step towards deep colonoscopic SLAM. Having successfully built an approach where mapping and localization work separately, we have paved the way to explore alternatives for a complete topological SLAM system able to consider relocalizations and loop closures while building the map.



\addtolength{\textheight}{-2cm}   


\printbibliography

\end{document}